\documentclass[runningheads]{llncs}
\usepackage[T1]{fontenc}
\PassOptionsToPackage{hyphens}{url}\usepackage[pdftex, colorlinks=true, hyperfootnotes=false, hyperindex=true, plainpages=false, pagebackref=false, pdfpagelabels=true, pdfstartview=FitH, linkcolor=blue, citecolor=blue, urlcolor=blue, bookmarks=false]{hyperref}
\usepackage{orcidlink}
\usepackage{graphicx}
\usepackage{microtype} 

\usepackage{enumitem} 
\usepackage{amsmath} 
\usepackage{graphicx}
\usepackage{subcaption}

\begin{document}
\title{David vs. Goliath in Next Activity Prediction:\\Argmax vs.\ LSTM, Transformer, and LLM}
\titlerunning{NAP: Argmax vs.\ LSTM, Transformer, and LLM}
%
%
\author{Hans Weytjens\inst{1}\orcidID{0000-0003-4985-0367} \and
Ingo Weber\inst{1,2}\orcidID{0000-0002-4833-5921} 
}
\authorrunning{H. Weytjens and I. Weber}
%
\institute{
Technical University of Munich, School of CIT, Germany, \and
    Fraunhofer Gesellschaft, Munich, Germany\\
    \email{firstname.lastname@tum.de}
}

\maketitle              
\begin{abstract}
Next activity prediction (NAP) is a cornerstone of predictive process monitoring (PPM), enabling organizations to move from retrospective analysis to proactive process steering. The PPM field has progressed from classical machine learning through deep learning architectures such as LSTMs and Transformers to large language models (LLMs).
Despite growing model complexity, no benchmark jointly compares LLMs, Transformers, LSTMs, and simple baselines in a direct sequence modeling setting for NAP. In this paper, we fill this gap with a systematic benchmark. We compare vocabulary-adapted LLMs, Transformers trained from scratch, LLM-distilled Transformers, and LSTMs against a simple counting-based argmax baseline across seven real-life event logs. Our results tell a David vs.\ Goliath story: pretraining confers no consistent improvement over training from scratch, model size shows little effect on performance, and on most datasets the argmax baseline matches or approaches the performance of billion-parameter LLMs.

\end{abstract}

\section{Introduction}\label{sec:introduction}
\vspace{-2mm}
Next activity prediction (NAP) constitutes one of predictive process monitoring (PPM)'s cornerstones. It enables organizations to move from retrospective analysis to proactive process steering. 

Initially, researchers turned to classical machine learning (ML) techniques such as decision trees, support vector machines, etc.\ for NAP tasks \cite{TAMA2019233}. As deep learning (DL) models became increasingly available, the field experimented with some multi-layer perceptrons (MLPs), before pivoting to convolutional neural networks (CNNs), and more importantly, recurrent neural networks (RNNs) with an emphasis on Long Short-Term Memory Networks (LSTMs) \cite{weinzierl2020empiricalcomparisondeepneuralnetworkarchitectures}. Later, some researchers followed up with Transformers \cite{wuyts2024sutran,bukhsh2021processtransformerpredictivebusinessprocess,nguyen2025next,van2023experiment}. The breakthrough of general-purpose large language models (LLMs) opened a new NAP research direction with text-serialization approaches, using in-context learning (ICL) \cite{rebmann2024evaluating,pasquadibisceglie2024lupin}, fine-tuning \cite{rebmann2024evaluating,CASCIANI2026102642,pyrih2025llms}, Retrieval-Augmented Generation (RAG) \cite{CASCIANI2026102642}, etc.

Oyamada et al. \cite{oyamada2025domainadaptationllmsprocess} applied a vocabulary-adapted direct sequence modeling approach (as used in classical ML, DL, and Transformers) to pretrained LLMs: They adapted the LLM's vocabulary by replacing the LLMs' input and output layers by custom embedding and linear layers to account for the NAP dataset's smaller vocabulary and different tokenization.  These customized LLMs were then fine-tuned using the NAP datasets. The goal was to ``evaluate the LLM's intrinsic capability to interpret sequential process information [...] bypassing the need for natural language conversion'' \cite{oyamada2025domainadaptationllmsprocess}.

In our work, we wish to test this hypothesis and establish to what extent the favorable comparison of the vocabulary-adapted, fine-tuned LLMs to LSTMs can be attributed to their pretraining. We also investigate how smaller pretrained LLMs compare with legacy GPT-2 and more recent Qwen-3 models. For lack of pretrained small LLMs, we use distillation to derive such models from larger teacher models. To relate to earlier research, we compare the Transformer-based models to LSTMs, and additionally, a simple count-based argmax model. The latter comparison led us to pose the provocative question of whether these advanced DL models are worth the effort.


Within the context of direct sequence modeling approaches for NAP, this paper makes the following contributions:
\begin{itemize}
    \item A systematic evaluation of the effect of generic LLM pretraining versus training from scratch.
    \item A distillation-based method to compress fine-tuned LLMs into smaller model architectures for which no pretrained versions exist.
    \item A comprehensive benchmark, comparing a spectrum of modern sequence modeling approaches— from LSTMs through Transformers to LLMs of varying scale and vintage.
    \item A sobering reality check: by including a count-based argmax baseline absent from prior work, we reveal the degree to which reported progress in NAP may have been overstated.
\end{itemize}

 In the following section, we provide background about NAP, LSTMs, Transformers, and LLMs and derive our research questions from it. Section~\ref{sec:4-methodology} explains the methodology and setup for our experiments. Next, we present the experimental results and answers to the research questions in Section~\ref{sec:6-Results}. We interpret these results in Section~\ref{sec:7-Discussion}, before concluding the paper in Section~\ref{sec:8-Conclusion}. The code for our work is available in our repository at \url{https://github.com/hansweytjens/DavidGoliath}.

\section{Background \& Research Questions}\label{sec:background}
\vspace{-2mm}

\subsection{Predictive Process Monitoring (PPM) and Next Activity Prediction (NAP)}
\vspace{-2mm}
PPM concerns predictions about the future of ongoing business process executions from their prefixes. In their systematic literature review of PPM methods, Di Francescomarino et al. \cite{10.1007/978-3-319-98648-7_27} identify three broad prediction categories: (1) numeric, such as time and cost predictions, (2) categorical, such as risk and categorical outcome predictions, and (3) next activities (plural) predictions. Within the latter, next activity (singular) prediction can be viewed as a specific case.

A process execution or case $c \in \mathcal{C}$ is represented by a trace $t_c = \langle e_1, e_2, \ldots, e_n \rangle$, where $e_i$ denotes the $i$-th event and $\mathcal{C}$ the set of all case identifiers. Each event $e_i$ consists of an activity label $a_i$, a timestamp $\tau_i$, and, optionally, further event attributes. Formally, $e_i = (a_i, \tau_i, \mathcal{A}_i)$ where $\mathcal{A}_i$ is the (possibly empty) set of additional attributes. An event log $\mathcal{L}$ is
a multiset of traces, i.e., $\mathcal{L} = \{ t_c \mid c \in \mathcal{C} \}$, with traces $t_c = \langle e_1, \ldots, e_n \rangle$, for which $\tau_i \leq \tau_{i+1} $. A trace's prefix of length $k$, with $1 \leq k \leq n$, is defined as $\operatorname{pref}_k(t_c) = \langle e_1, \ldots, e_k \rangle$. The set of all prefixes is therefore $\mathcal{P} = \{ \operatorname{pref}_k(t_c) \mid t_c \in \mathcal{L},\ 1 \leq k \leq |t_c| \}$. Given a running case with an observed prefix $\sigma = \operatorname{pref}_m(t_c)$, with $\ 1 \leq m < |t_c|$, next activity prediction aims to predict the activity label $a_{m+1}$ of the next event $e_{m+1}$.

\vspace{-1mm}
\subsection{Long Short-Term Memory Networks (LSTMs) for NAP}
\vspace{-2mm}
In 1997, Hochreiter et al. \cite{10.1162/neco.1997.9.8.1735} proposed the LSTM, an enhancement of the RNN, particularly suited to capture long-term dependencies in sequential data. Two decades later, Evermann et al. \cite{EVERMANN2017129} and Tax et al. \cite{10.1007/978-3-319-59536-8_30} applied LSTMs to NAP tasks. LSTMs emerged as a strong and widely adopted technique for NAP, often outperforming classical machine learning approaches \cite{10.1007/978-3-319-59536-8_30}, even though they are not universally superior to all alternative neural architectures \cite{rama2021deep}. However, the sequential nature of LSTMs limits parallelization and requires information to be propagated across many time steps, making the modeling of very long-range dependencies in event log sequences challenging. These limitations motivated the adoption of attention-based architectures such as the Transformer.

\vspace{-1mm}
\subsection{Transformers and Large Language Models}\label{subsec:Transf_LLM}
\vspace{-2mm}
Twenty years after the LSTM, Vaswani et al. \cite{vaswani2017attention} proposed the Transformer, originally formulated as an encoder–decoder architecture based on attention mechanisms. The following year, Radford et al. \cite{radford2018improving} adopted a decoder-only Transformer and demonstrated that large-scale self-supervised pretraining with a language modeling objective, followed by task-specific fine-tuning, yields strong performance across diverse downstream tasks. This model family, termed the Generative Pre-trained Transformer (GPT), forms the basis of many modern LLMs — a term that was adopted as model capacity and training data scale increased.

In practice, when applying Transformer architectures and LLMs to dataset-driven tasks (rather than conversational settings), several modes of use can be distinguished. First, a Transformer model can be trained from scratch, i.e., initialized with random weights and optimized directly on the target dataset, analogous to conventional DL models. Second, with in-context learning (ICL) \cite{brown2020language}, a pretrained LLM is used without modifying its parameters by providing a prompt that may contain task instructions, contextual information, and data ranging from a few representative examples to an entire dataset in structured form. Third, the model can be fine-tuned \cite{radford2018improving}, where the pretrained weights are further adapted to a specific domain or task using supervised or self-supervised objectives on task-specific data. Finally, knowledge distillation \cite{hinton2015distilling} can be applied, where a smaller “student” model is trained on task-specific data using supervision derived from a larger pretrained “teacher” model (e.g., soft labels or logits), enabling more efficient deployment while retaining much of the teacher’s performance.

\vspace{-1mm}
\subsection{Transformers and LLMs in NAP}
\vspace{-2mm}
We distinguish two main approaches to using Transformers and LLMs for NAP: namely text-serialization approaches, which convert event data to natural language, and direct sequence modeling approaches, which preserve the structured nature of event data by representing event attributes directly as learned embeddings, without natural language conversion. Our work tackles the latter category.

\vspace{-1mm}\paragraph{1) Text-serialization approaches} aim to leverage the contextual reasoning capabilities of LLMs by transforming structured event log data into textual encodings that can be processed by LLMs.

Rebmann et al. \cite{rebmann2024evaluating} found NAP too challenging for in-context learning (ICL) with off-the-shelf LLMs but reported improved performance after task-specific fine-tuning. As an alternative to both ICL and fine-tuning, Casciani et al. \cite{CASCIANI2026102642} rely on Retrieval-Augmented Generation (RAG) to provide selected event data as additional context to the model. Rather than fine-tuning on a single task such as NAP, Pyrih et al. \cite{pyrih2025llms} adopt instruction-tuning across multiple downstream tasks, achieving mixed results with modest improvements for NAP compared to earlier findings \cite{rebmann2024evaluating}. Further variations include fine-tuning LLMs directly on serialized XES logs \cite{10.1007/978-3-031-78666-2_15},  and training encoder-only models for suffix prediction \cite{pasquadibisceglie2024lupin}. The text-serialization approach permits fine-tuning LLMs on multiple event log datasets from different domains \cite{rebmann2024evaluating,pyrih2025llms}, effectively realizing transfer training. However, since some publicly available datasets may have been included in LLM pretraining corpora, data contamination poses a potential threat to the validity of empirical evaluations.

\vspace{-1mm}\paragraph{2) Direct sequence modeling} has two variants:

\textbf{Training Transformers from scratch:} 
Wuyts et al. \cite{wuyts2024sutran} trained a randomly-initialized encoder-decoder Transformer for suffix prediction. Later, Bukhsh et al. \cite{bukhsh2021processtransformerpredictivebusinessprocess} opted for a decoder-only variant, claiming superior NAP performance compared to earlier methods including LSTM. Higher accuracy rates were claimed by adding a mixture-of-experts (MoE) mechanism \cite{nguyen2025next} and transfer learning \cite{van2023experiment}.  Berti et al. \cite{berti2026context} also use an MoE and train an ``in-context foundation model'' using several synthetic datasets at once.

\textbf{Vocabulary Adaptation:} is a transfer learning strategy in which a pretrained LLM backbone is reused while a corpus-specific vocabulary and corresponding tokenization scheme is defined a priori for the target dataset. This induces a new input embedding matrix and output projection that are trained to map the new tokens into and out of the pretrained model’s latent space, whereas the core Transformer backbone remains fixed or only minimally adapted. De Vries et al. \cite{de2021good} explored this idea in the context of cross-lingual transfer. Oyamada et al. \cite{oyamada2025domainadaptationllmsprocess} extended the approach to the PPM domain for NAP and remaining time predictions, applying various parameter-efficient fine-tuning (PEFT) techniques.

\vspace{-1mm}
\subsection{Research Questions}
\vspace{-2mm}
Against this backdrop, we are now ready to formulate our research questions.
In this paper, we investigate the added value of generic next-token pretraining in LLMs for NAP using vocabulary adaptation and systematically comparing pretrained LLM backbones to Transformers trained from scratch as well as to smaller distilled models. To structure this investigation, we pose the following research questions:

\begin{enumerate}[label=RQ\arabic*:, leftmargin=*]
    \item Does generic LLM pretraining improve NAP performance?
    \item Can knowledge distillation from fine-tuned teacher models improve the performance of smaller architectures?
    \item Do larger models outperform smaller ones?
    \item Do more recent LLM architectures (e.g., Qwen-3 from 2025) outperform earlier ones (e.g., GPT-2 from 2019)?
    \item Does the high complexity of (fine-tuned) LLMs and Transformers offer performance gains over LSTMs and an argmax baseline?
\end{enumerate}

\vspace{-1mm}
\section{Methodology and Experimental Setup}\label{sec:4-methodology}
\vspace{-2mm}
\subsection{Models and baseline}
\vspace{-2mm}
Table~\ref{models} lists the models\footnote{Model sizes differ from the official numbers because we modified the vocabulary, which changes the input and output layers. Since each dataset has a different vocabulary, model sizes vary slightly across datasets; the reported values correspond to BPI12. Architectural differences between GPT-2 and nanoGPT also lead to different parameter counts.} used in our experiments. We selected three models from OpenAI's open source GPT-2
\footnote{Hugging Face GPT-2 checkpoints:
\texttt{openai-community/gpt2},
\texttt{gpt2-medium}, and
\texttt{gpt2-large},
(\url{https://huggingface.co/openai-community}, accessed 2026-03-10).}\footnote{To avoid confusion, we refer to ``GPT-2'' as ``GPT-2-small''.} LLM family enabling controlled scaling comparisons across model sizes under a shared architecture, training corpus (OpenWebText, $\approx 40$\,GB), vocabulary, and training procedure. To this, we added GPT-2-mini\footnote{\url{https://huggingface.co/erwanf/gpt2-mini}, accessed 2026-03-10. }, a similar but smaller model trained on a subset of OpenWebText. The Qwen-3 LLM series\footnote{\url{https://huggingface.co/collections/Qwen/qwen3}, \url{https://qwen.ai/blog?id=qwen3}, accessed 2026-03-10.} adds more modern and larger GPT-style models to our selection, whilst retaining some capacity overlap with the larger GPT-2 models. Our choice for GPT-2 was also motivated by Andrej Karpathy's nanoGPT, a minimal GPT-2–style Transformer implementation\footnote{\url{https://github.com/karpathy/nanochat}, accessed 2026-03-10.} which we instantiate in different sizes, reflected in their names in Table~\ref{models}. We use nanoGPT for from-scratch training and distillation. To compare with a classical ML model, we also included four LSTM models of different sizes. 

Finally, we compare these models with an \textbf{argmax} baseline. Our David is a first-order empirical transition model that predicts the next activity as $a_{k+1} = \arg\max_{a} P(a \mid a_k)$.
$P(a \mid a_k)$ is estimated from the relative transition frequencies in the training set. It conditions only on the current activity and requires no optimization.

\begin{table}[tb]
\centering
\caption{Models. Sizes refer to values for BPI12.}\label{models}
\begin{tabular}{|l|c|c|c|c|}
\hline
Model & Total & Hidden & Layers & Att. heads\\
 & Params & (embedding) size &  & \\
\hline
{\bfseries GPT-2} &&&&\\
GPT-2-large &774\,M&1,280&36&20\\
GPT-2-medium &354\,M&1,024&24&16\\
GPT-2-small &124\,M&768&12&12\\
GPT-2-mini (not OpenAI) &39\,M&512&4&8\\

\hline
{\bfseries Qwen-3} &&&& (Q / KV)\\
Qwen-3-4B &4\,B &2,560&36&32/8\\
Qwen-3-1.7B &1.7\,B&2,048&28&16/8\\
Qwen-3-0.6B &596\,M&1,024&28&16/8\\
\hline
{\bfseries nanoGPT} &&&&\\
Nano-768-12-12 & 85\,M  & 768 & 12 & 12 \\
Nano-512-12-8  & 38\,M  & 512 & 12 &  8 \\
Nano-256-12-4  & 9.5\,M  & 256 & 12 &  4 \\
Nano-128-12-2  & 2.4\,M  & 128 & 12 &  2 \\
Nano-768-4-12  & 28\,M  & 768 &  4 & 12 \\
Nano-512-4-8   & 13\,M  & 512 &  4 &  8 \\
Nano-256-4-4   & 3.2\,M   & 256 &  4 &  4 \\
Nano-128-4-2   & 0.8\,M   & 128 &  4 &  2 \\
Nano-64-4-1    & 0.2\,M  &  64 &  4 &  1 \\
Nano-64-2-1    & 0.1\,M  &  64 &  2 &  1 \\
\hline
{\bfseries LSTM} &&&&\\
LSTM-256-2 &1.1\,M&256&2&-\\
LSTM-128-2 &0.3\,M&128&2&-\\
LSTM-64-2 &0.07\,M&64&2&-\\
LSTM-64-1 &0.04\,M&64&1&-\\
\hline
{\bfseries Baseline} &&&&\\
argmax &-&-&-&-\\
\hline
\end{tabular}
\end{table}

\vspace{-1mm}
\subsection{Datasets and preprocessing}\label{subsec:datasets}
\vspace{-2mm}
We use the BPI challenge event logs\footnote{\url{https://data.4tu.nl/}, accessed 2026-03-10.} BPI12, BPI15, BPI17, BPI19, Prepaid Travel Costs (BPI20\_PTC), Travel Permit Data (BPI20\_TPD), and Request for payment (BPI20\_RFP)\footnote{These are the same datasets as Oyamada et al. \cite{oyamada2025domainadaptationllmsprocess}, with the exception of BPI15. All other cited papers using real-life datasets \cite{EVERMANN2017129,10.1007/978-3-319-59536-8_30,wuyts2024sutran,rama2021deep,bukhsh2021processtransformerpredictivebusinessprocess,nguyen2025next,van2023experiment,CASCIANI2026102642,10.1007/978-3-031-78666-2_15,pasquadibisceglie2024lupin} relied at least partially on some of these as well.}. Table~\ref{datasets} demonstrates how these datasets vary along multiple dimensions affecting NAP difficulty. A high share of top-5 and top-10 prefixes (BPI19: 0.42/0.59, BPI20\_RFP: 0.58/0.78) means that a small number of distinct prefixes occur very frequently, indicating strong concentration in the prefix distribution. A low weighted conditional entropy\footnote{$ H = - \sum_{\sigma \in \mathcal{P}} P(\sigma) \sum_{a \in \mathcal{A}} P(a \mid \sigma) \log_2 P(a \mid \sigma) $} (BPI15: 0.25, BPI17: 0.32) indicates that, given a prefix, the next activity is highly predictable. Datasets combining frequent prefixes with few classes and low conditional entropy (BPI20\_RFP, BPI20\_PTC) are expected to favor the argmax baseline. BPI15 — with 168K unique prefixes out of 201K and 324 classes — presents the opposite case, despite its low entropy. The performance of more sophisticated models also depends on the availability of explanatory variables in the dataset (unknown) and the number (known) and quality (unknown) of samples to learn from.

We reject traces with fewer than three events. For greater relevance, we incorporate lifecycle information by augmenting activity labels where available. We attach an ``EOS'' token to each case because case completion is itself a relevant next-activity prediction outcome. We do not cap the number of classes to observe the detrimental effect of high numbers (see BPI15). To avoid data leakage and test set bias, we build train and test sets (20\%) according to Weytjens et al. \cite{10.1007/978-3-030-94343-1_2}. A validation set is then extracted from the training set. We calculate the ``accumulated\_time'' (since the case's first event's timestamp) attribute for each event.

\begin{table}[tb] 
\centering 
\caption{Datasets after preprocessing (as per Section~\ref{subsec:datasets}).}\label{datasets} 
\begin{tabular}{|l|c|c|c|c|c|c|c|} 
\hline Dataset & No. of & No. of & No. of& No. of & Top 5 & Top 10  & Weighted \\
& cases & classes & prefixes & unique & prefixes & prefixes   & conditional \\
& & & & prefixes &(\%) &(\%)&  entropy \\
\hline BPI12& 9,487 & 37& 173,498 & 37.060 & 16.93 & 24.36 &  0.40 \\
BPI15& 4,411 & 324& 201,243 & 168,344 & 3.70 & 4.65 &  0.25 \\
BPI17& 29,306 & 63 & 1,100,691 & 246,844 & 9.22 &13.98 &  0.32 \\
BPI19& 156,072& 41& 914,988 & 40,058 & 41.67 & 59.03 &  0.75 \\
BPI20\_PTC& 1,791 & 30 & 15,730 & 940 & 30.03 & 44.69 & 0.47 \\
BPI20\_TPD & 6,847& 52 & 83,620 & 13,134 & 20.76 & 28.04 &  0.58 \\
BPI20\_RFP& 5,701 & 18 & 30,153 & 245 & 57.57 & 78.39 & 0.49 \\ 
\hline 
\end{tabular} 
\end{table}

\vspace{-1mm}
\subsection{Training}
\vspace{-2mm}

\vspace{-1mm}\paragraph{Data representation and model adaptation:} 

Preliminary test runs using the validation set (see our code repository) led to the selection of the following attributes\footnote{The attribute names are standardized; original names differ across datasets.} for training: ``activity'' (all datasets), ``resource'' (all), ``amount'' (BPI12, BPI17), and ``accumulated\_time'' (all). The numerical features are standardized. The categorical feature ``activity\_name'' is concatenated with the ``lifecycle'' information (e.g., ``start'', ``complete'') when available. Categorical values are mapped from strings to integer IDs using the training dataset vocabulary (with special PAD/UNK/EOS tokens). Each ID is mapped via a learned embedding table to an $E$-dimensional vector (with $E$ equal to the model's embedding/hidden size). In parallel, the numerical feature is projected to the same $E$-dimensional space through a linear layer. The categorical and numerical representations are then combined by element-wise addition (``sum'' strategy) before being passed to the backbone. Prefixes are variable-length event sequences and are right-padded with PAD tokens to the longest trace in each mini-batch; an attention mask is applied to ensure that padded positions are ignored by the attention mechanism. The effective maximum sequence length is bounded by the backbone context window.

\vspace{-1mm}\paragraph{General:} Following the same procedure~\cite{10.1007/978-3-030-94343-1_2} as for the test set, we carve validation sets (10\%) out of the training sets. We minimize the cross-entropy loss with the Adam optimizer. Batch sizes are 64. We do not regularize the models, but enforce early stopping (patience = 10) to combat overfitting. We conducted preliminary pilot experiments to select appropriate learning rates. Based on these experiments, a learning rate of 0.005 was used for nanoGPT and smaller GPT-2 models (up to GPT-2-medium), while 0.0005 was used for larger models. Learning-rate tuning results are available in our code repository.

\vspace{-1mm}\paragraph{Fine-tuning:} We replaced the input and output layers of the pretrained LLMs to map embeddings and logits to our dataset-specific vocabulary, applying weight-tying \cite{press2017using} between the input embedding and output projection matrices, while keeping the Transformer backbone architecture intact. We used parameter-efficient fine-tuning (PEFT \cite{han2024parameter}), exploring selective layer freezing strategies \cite{app151910434}. Given our vocabulary adaptation of both input and output layers, we considered unfreezing both bottom and top layers. Our experiments showed that freezing all blocks except the first and last, with LayerNorm kept trainable, consistently matched or outperformed other configurations including full fine-tuning. Ablation results are available in our code repository.

\vspace{-1mm}\paragraph{Distillation:}
We use logits-only knowledge distillation, where the student is trained to match the teacher’s output distribution directly. This design is architecture-agnostic and requires no intermediate feature alignment. The training objective $\mathcal{L} = (1-\alpha)\mathcal{L}_{CE} + \alpha\mathcal{L}_{KL}$ combines hard-label cross-entropy and KL divergence to teacher logits, with $\alpha$ increased from 0.1 to 0.9 over training. For each dataset, the fine-tuned model (GPT-2-x or Qwen-3-x) with the lowest validation loss assumes the role of teacher, with all Nano-x-x-x models its students.

\vspace{-1mm}\paragraph{Hardware and framework:}
We used NVIDIA A100 GPUs and a NVIDIA GB10 Grace-Blackwell superchip, depending on availability. All LLMs were sourced from Hugging Face and run through the Transformers library. 

\vspace{-1mm}
\subsection{Evaluation Metrics}
\vspace{-2mm}
We evaluate model performance using two complementary metrics: accuracy
and macro-F1, both commonly used in the next-activity prediction literature.

Accuracy measures the proportion of prefixes for which the correct next activity is predicted. However, the datasets exhibit substantial class imbalance, as indicated by the prefix
concentration in Table~\ref{datasets}. In such settings, predicting
only the dominant activity may yield high accuracy while failing on rare activities.
We therefore additionally report macro-F1, the harmonic mean of recall and precision, which assigns equal weight to each activity
class regardless of frequency. Metrics are computed at the prefix level, where each prefix corresponds to one prediction
instance in the test set.

Rather than reporting the full set of statistical significance tests inline, we provide Wilcoxon signed-rank test statistics and p-values in the code repository. Figures~\ref{fig:pretraining}--\ref{fig:LSTM_baseline} include standard deviations to support visual interpretation.

\vspace{-1mm}
\section{Results}\label{sec:6-Results}
\vspace{-2mm}

\begin{figure}[p]
\centering
\includegraphics[width=.85\textwidth]{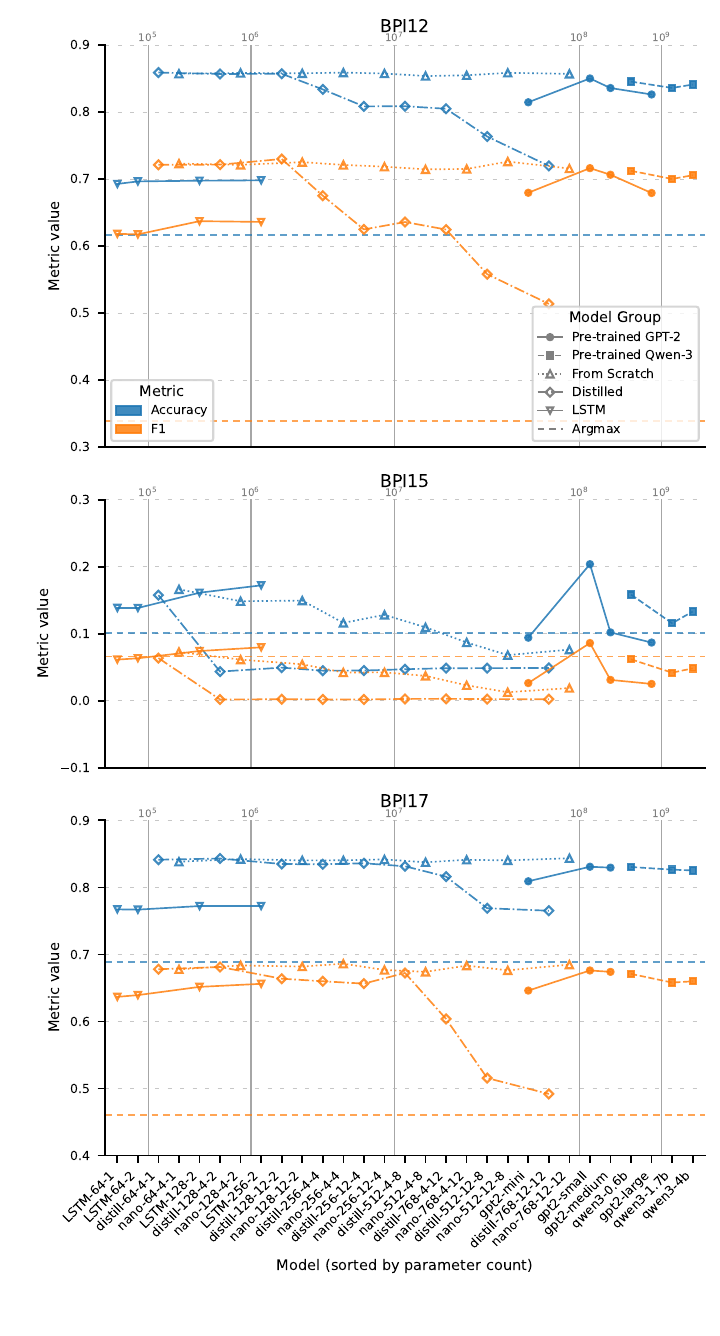}
\caption{Accuracy and macro-F1 as a function of total number of parameters. All models are displayed at equidistance, with vertical grid lines showing approximate size on a log scale. The vertical axes differ per dataset, but are at the same scale for better comparison. Standard deviations are omitted for clarity but are reported in tables in the code repository (see Section~\ref{sec:introduction}).} \label{fig:alltogehter_1}
\end{figure}

\begin{figure}[p]
\centering
\includegraphics[width=.85\textwidth]{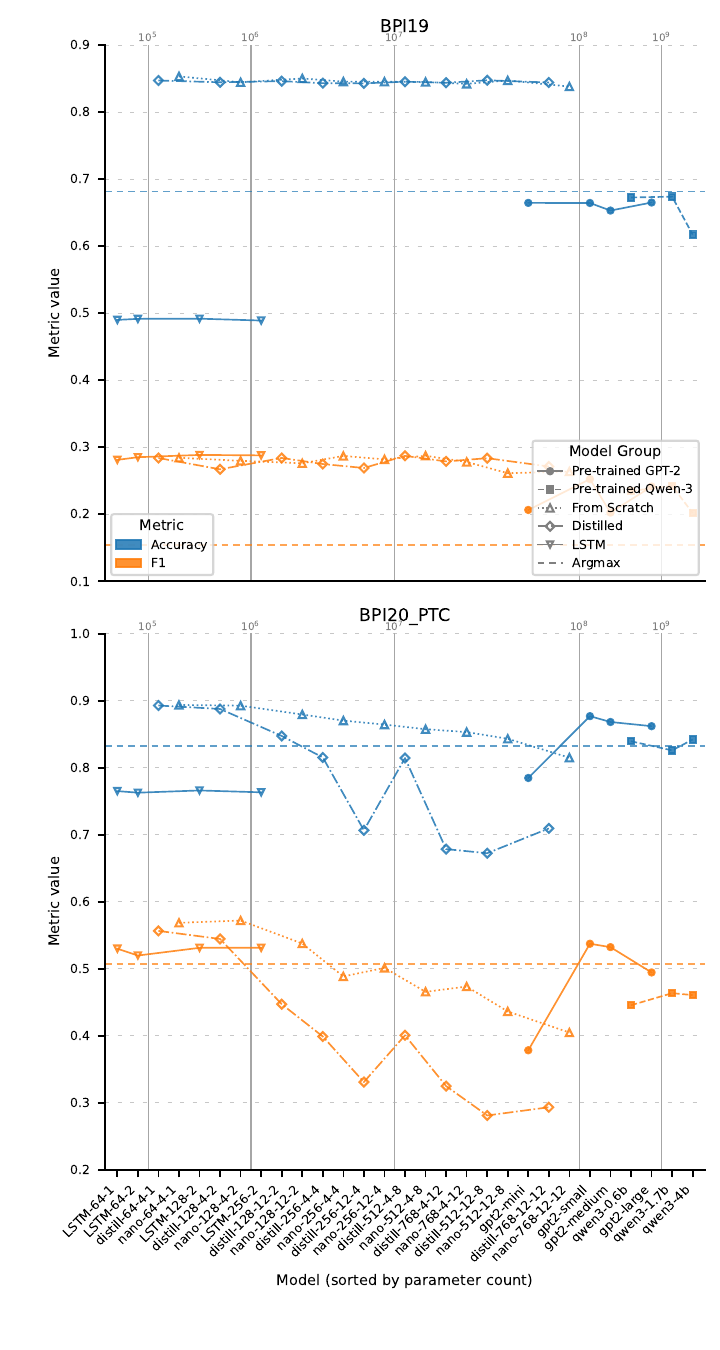}
\caption{Accuracy and macro-F1 as a function of total number of trainable parameters. Continuation of graphs on previous page.} \label{fig:alltogehter_2}
\end{figure}

\begin{figure}[p]
\centering
\includegraphics[width=.85\textwidth]{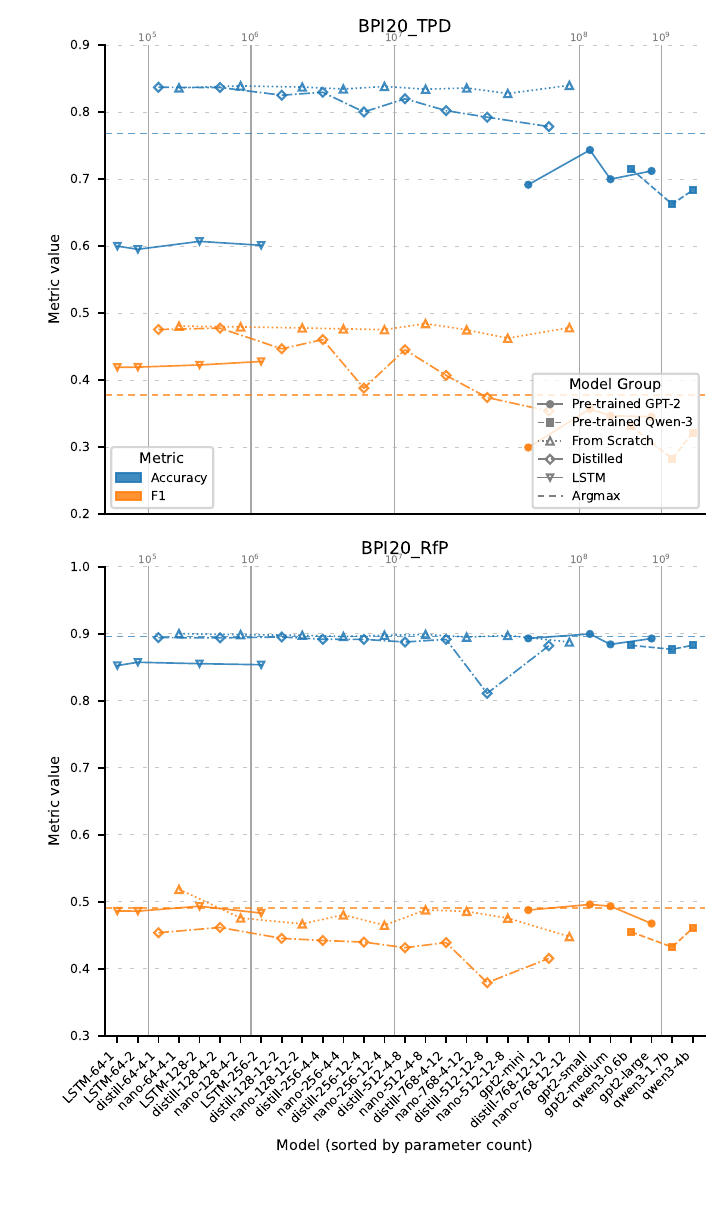}
\caption{Accuracy and macro-F1 as a function of total number of trainable parameters. Continuation of graphs on two previous pages.} \label{fig:alltogehter_3}
\end{figure}

Figures~\ref{fig:alltogehter_1}--\ref{fig:alltogehter_3} present the complete results; the figures in the remainder of this section highlight cross-sections corresponding to each research question. Preliminary hyperparameter search results (attribute selection, layer freezing, learning rates) are available in our code repository (see Section~\ref{sec:introduction}).



\subsection{Pretraining effect: direct measurement (RQ1)}
\begin{figure}[tb]
    \centering
    \begin{subfigure}[b]{0.49\textwidth}
        \centering
        \includegraphics[width=\textwidth]{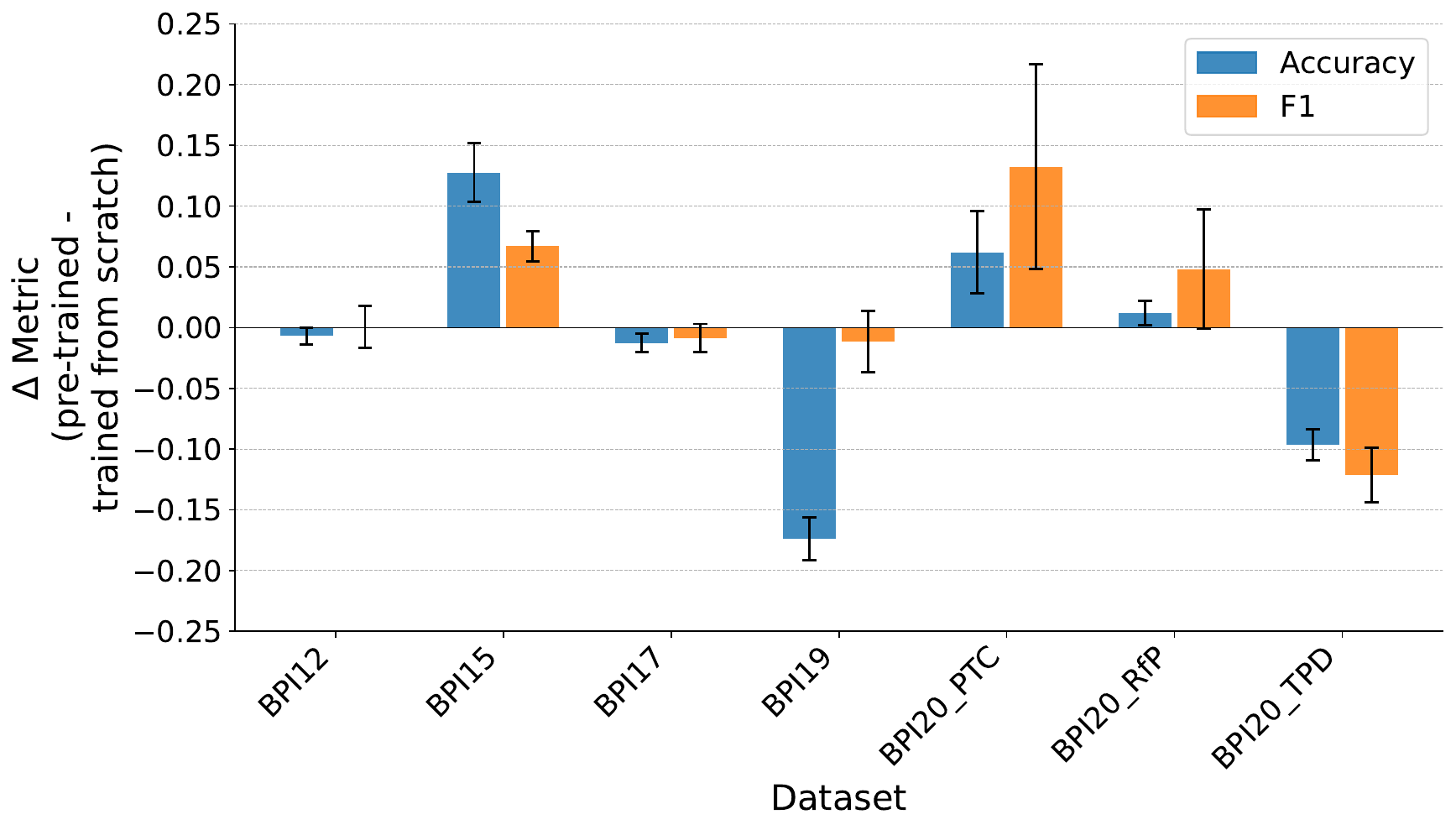}
        \caption{GPT-2-mini vs. Nano\_512\_4\_8.}
        \label{fig:image1}
    \end{subfigure}
    \hfill
    \begin{subfigure}[b]{0.49\textwidth}
        \centering
        \includegraphics[width=\textwidth]{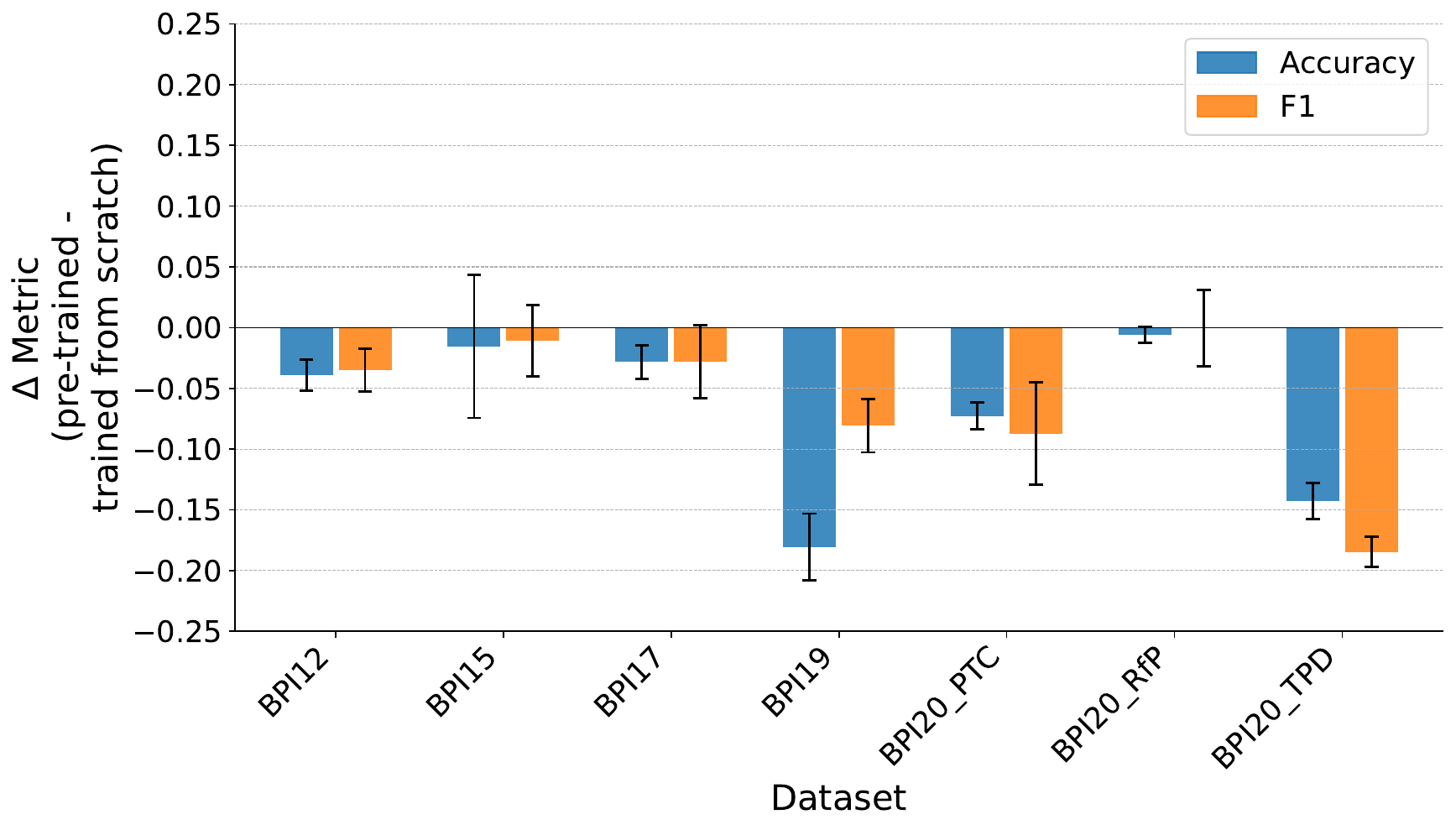}
        \caption{GPT-2-small vs. Nano\_768\_12\_12.}
        \label{fig:image2}
    \end{subfigure}
    \caption{Net pretraining effect for two model sizes.}
    \label{fig:pretraining}
\vspace{2mm}
\end{figure}

\vspace{-1mm}
In Fig.~\ref{fig:pretraining}, we compare the fine-tuned GPT-2-mini and GPT-2-small to their corresponding nanoGPT versions (Nano-512-4-8 and Nano-768-12-12), trained from scratch. No consistent pattern across datasets or model sizes emerges. Performance differences vary in magnitude and direction, indicating that pretraining yields no systematic advantage.

\vspace{1mm}
\noindent\textbf{RQ1 Result:} We observe no consistent added value of pretraining.

\vspace{-1mm}
\subsection{Pretraining effect: measurement with distillation (RQ2)}
\begin{figure}[tb]
    \centering
    \begin{subfigure}[b]{0.49\textwidth}
        \centering
        \includegraphics[width=\textwidth]{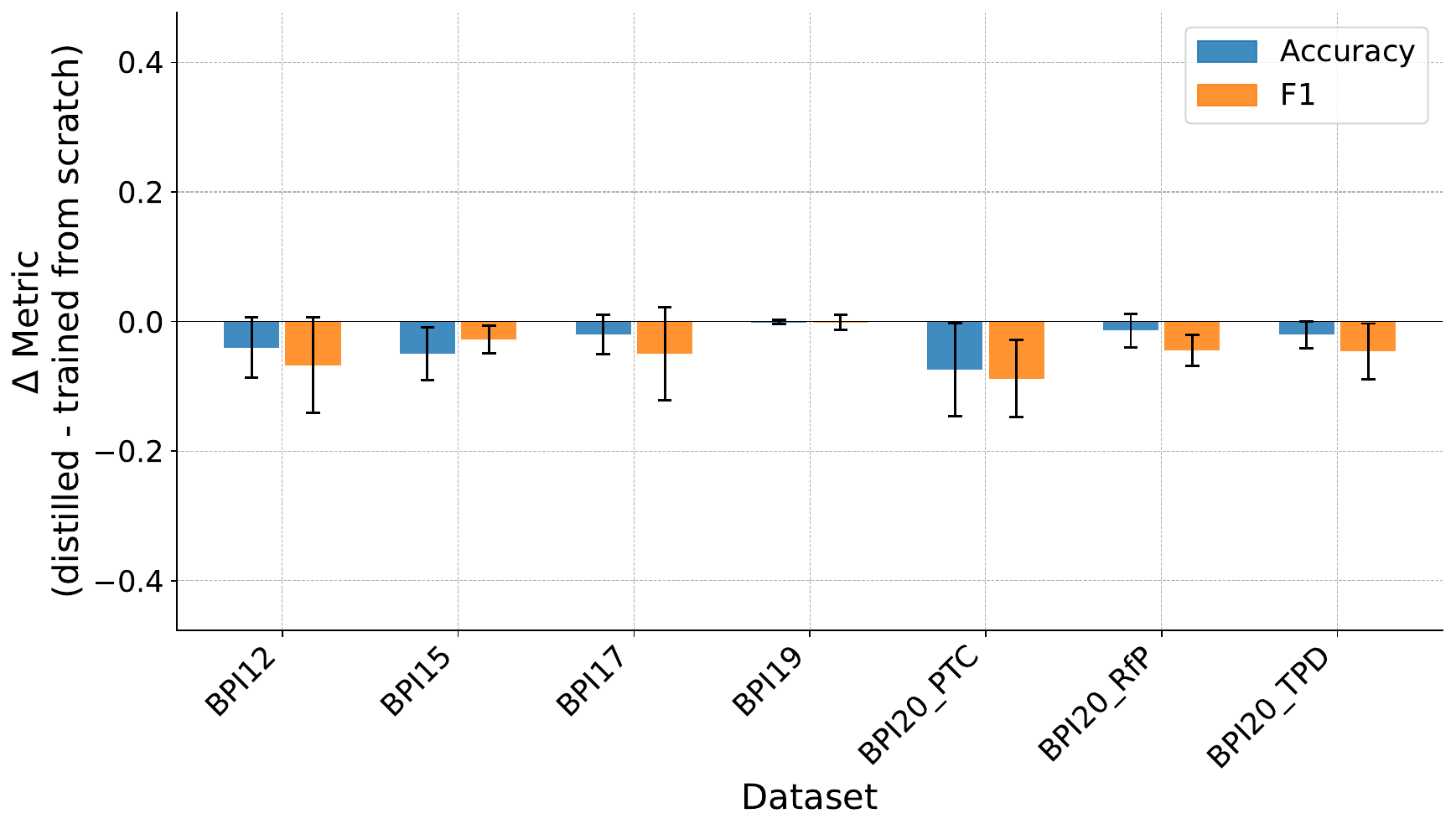}
        \caption{Per dataset, averaged over models.}
        \label{fig:dist11}
    \end{subfigure}
    \hfill
    \begin{subfigure}[b]{0.49\textwidth}
        \centering
        \includegraphics[width=\textwidth]{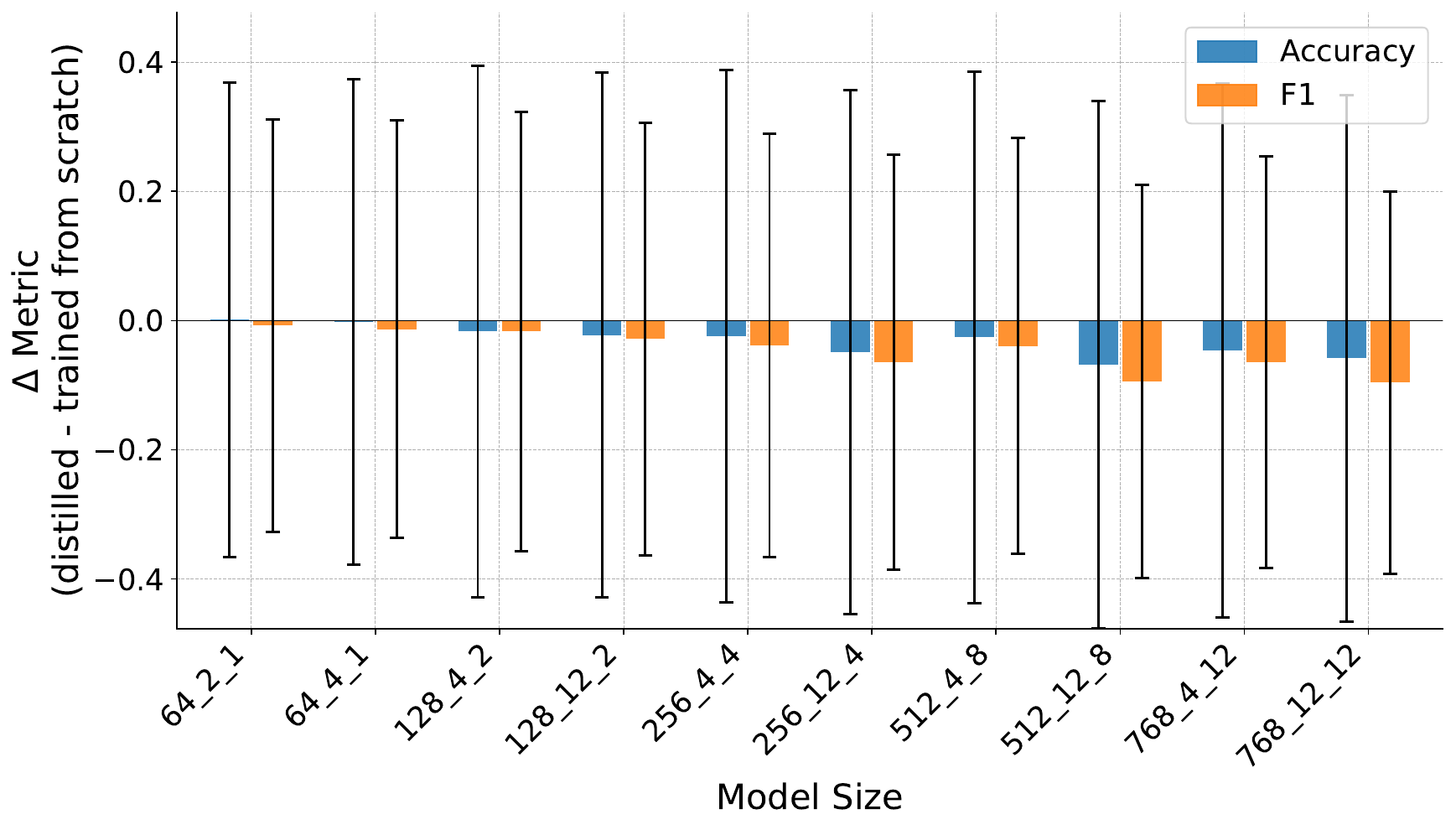}
        \caption{Per model, averaged over datasets.}
        \label{fig:dist12}
    \end{subfigure}
    \caption{Performance difference between distilled models and their corresponding models trained from scratch. Negative values indicate an advantage for training from scratch.}
    \label{fig:distillation}
    \vspace{2mm}
\end{figure}

\vspace{-2mm}
To investigate whether the observed phenomenon holds for smaller model architectures for which we have no pretrained version, we use distillation to ``create'' smaller fine-tuned models: the distilled student model learns concurrently from both the training data and the task-specific output distribution of its pretrained and fine-tuned teacher (see Section~\ref{subsec:Transf_LLM}). The student thus inherits the teacher's task-specific distribution, itself a product of its pretraining, carrying richer supervision than hard labels alone \cite{hinton2015distilling}. We train both from scratch and distill all remaining nanoGPT models. Across model sizes in Figure~\ref{fig:distillation}, average performance differences between distilled models and their corresponding
models trained from scratch are mostly negative despite variation across datasets, suggesting that distillation provides no systematic benefit. 

Closer inspection of Figures~\ref{fig:alltogehter_1}--\ref{fig:alltogehter_3} reveals that the smallest models trained from scratch score higher on both metrics than the much larger fine-tuned GPT-2 and Qwen-3 models, further undermining the case for using pretrained models.

\vspace{1mm}
\noindent\textbf{RQ2 Result:} Logits-only distillation from fine-tuned pretrained models does not improve smaller architectures over training them directly from scratch.

\vspace{-1mm}
\subsection{Model capacity and modernity (RQ3 and RQ4)}
\vspace{-2mm}
Figures~\ref{fig:alltogehter_1}--\ref{fig:alltogehter_3} show that for most model families (nanoGPT, GPT-2, Qwen-3) results are very similar across sizes within the respective families. The early stopping mechanism effectively prevents overfitting during training. The family of distilled models is the exception: performance \textit{degrades} with model size.

\vspace{1mm}
\noindent\textbf{RQ3 Result:} Within the range tested, increasing model size confers no consistent performance benefit. 

\vspace{5mm}
For similar sizes, the newer Qwen-3 models outperform the GPT-2 models slightly for BPI12 and  BPI15, but lag behind for the BPI20 datasets. The picture is unclear for BPI17 and  BPI19. In summary, our limited sample suggests no advantage for newer models.

\vspace{1mm}
\noindent\textbf{RQ4 Result:} The more recent Qwen-3 architecture shows no consistent advantage over GPT-2.

\vspace{-1mm}
\subsection{LSTMs and baseline (RQ5)}
\begin{figure}[tb]
    \centering
    \begin{subfigure}[b]{0.49\textwidth}
        \centering
        \includegraphics[width=\textwidth]{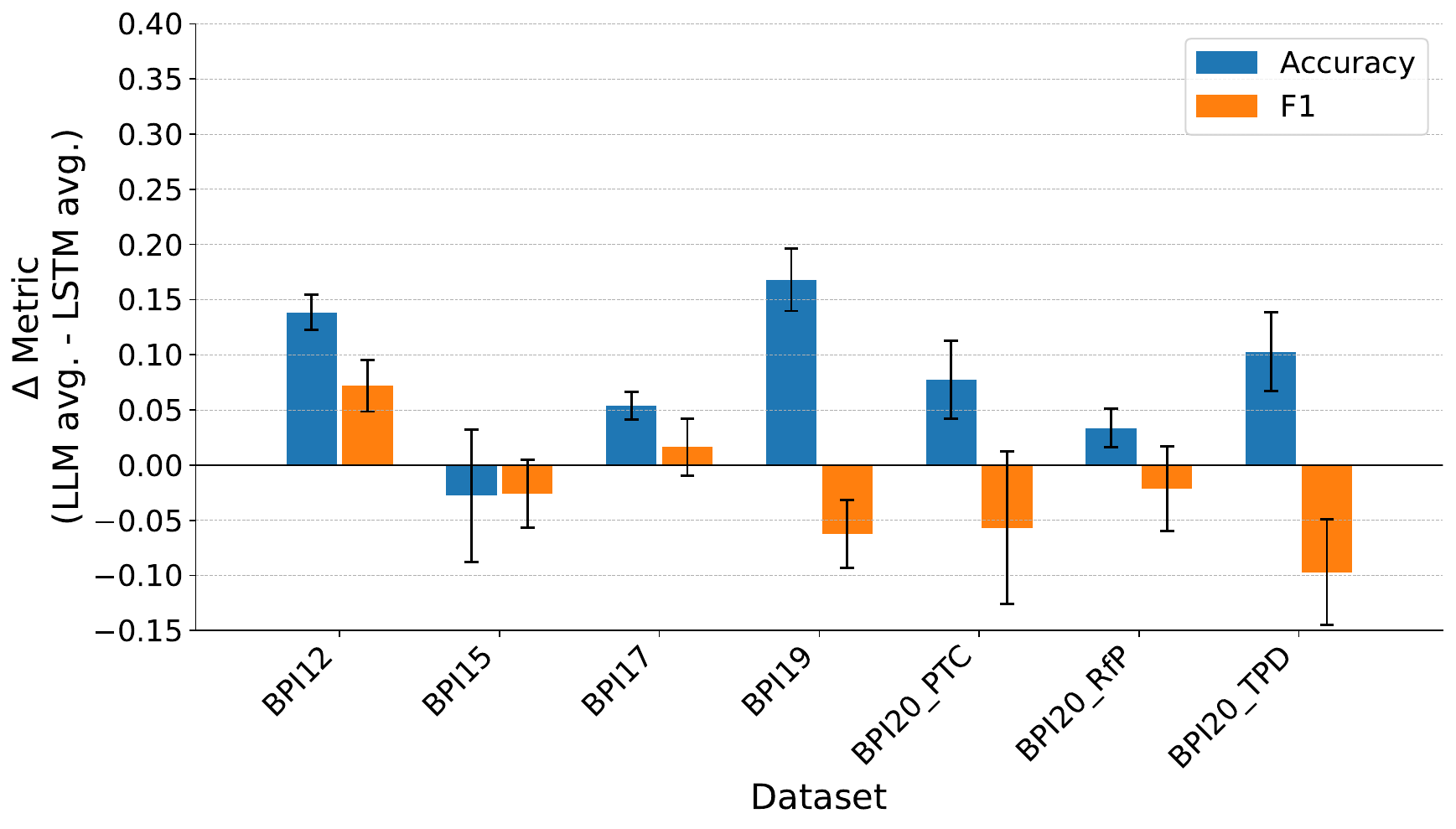}
        \caption{LLM vs. LSTM}
        \label{fig:dist21}
    \end{subfigure}
    \hfill
    \begin{subfigure}[b]{0.49\textwidth}
        \centering
        \includegraphics[width=\textwidth]{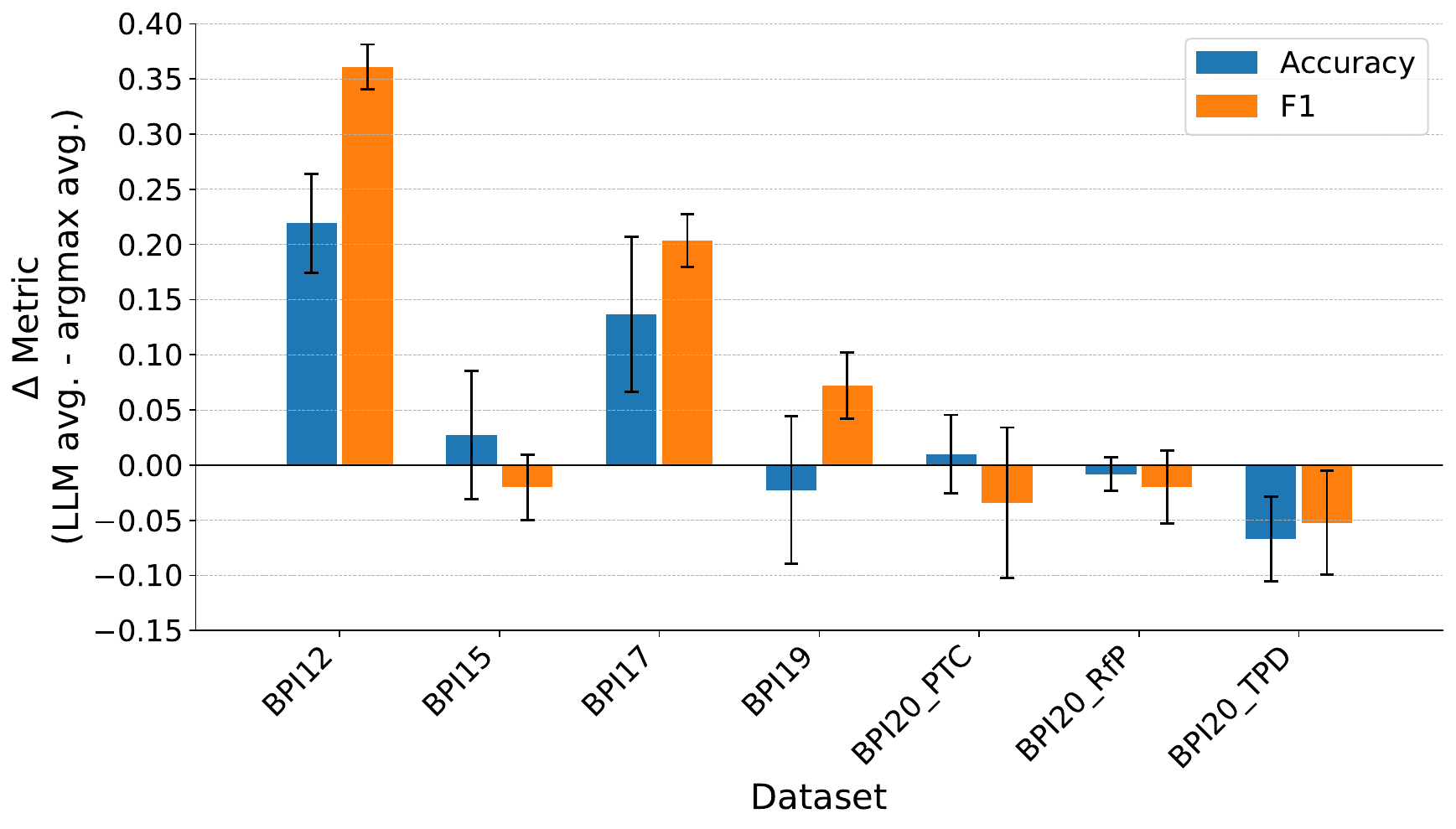}
        \caption{LLM vs. baseline}
        \label{fig:dist22}
    \end{subfigure}
    \caption{LLMs vs. LSTM and baseline per log. Averaged over all models within the family (fine-tuned GPT-2 and Qwen-3).}
    \label{fig:LSTM_baseline}
     \vspace{2mm}
\end{figure}

\vspace{-2mm}
Overall, the LLMs (fine-tuned GPT-2 and Qwen-3) reach higher accuracy than the LSTMs, as shown in the left panel of Figure~\ref{fig:LSTM_baseline}. Interestingly, the reverse is true for the macro-F1, where the LLMs only score significantly better on the BPI12 dataset. The picture for the LLMs further darkens when comparing them against the argmax baseline (right panel): only for the BPI12 and BPI17 datasets, the LLMs outperform argmax.

\vspace{1mm}
\noindent\textbf{RQ5 Result:} LSTMs come close to the LLMs' performance when considering both metrics. But for datasets with few training samples, high prefix variance, insufficient explanatory variables, and/or many different activities, defaulting to an argmax model is the practitioner's best option.

\vspace{-1mm}
\subsection{Wall-clock time}
\vspace{-2mm}
As modern GPUs are fast and relatively affordable, wall-clock time is of secondary importance compared to accuracy and F1. On our hardware, times range from 1--2 seconds for argmax, 0.5--11 minutes for LSTM, 0.6--15 minutes for nanoGPT models, 0.4--38 minutes for the distilled models, 2--56 minutes for GPT-2, and 4--281 minutes for Qwen-3. More details can be found in our code repository (see Section~\ref{sec:introduction}).



\vspace{-1mm}
\section{Discussion}\label{sec:7-Discussion}
\vspace{-1mm}
 Overall, the results indicate limited benefits of using pretraining and model scaling in our setting. A simple frequency-based baseline can be a competitive option, although comparatively small Transformer models trained from scratch achieve the best predictive performance on several datasets. 

\vspace{-1mm}\paragraph{Datasets}
Across datasets, the ranking of models by accuracy correlates highly with rankings by F1 (except for LSTM, see RQ5), validating our conclusions across both metrics. Only BPI12 and BPI17 yield meaningful improvements over the argmax baseline. The F1 gains exceed the accuracy gains, suggesting that learned models handle minority activity classes better than the argmax baseline. BPI15 proves intractable: 324 activity classes overwhelm all models. The high accuracy but very low F1 for BPI19 reflects the dataset's class imbalance: models learn the dominant classes while ignoring rare ones. The high argmax baseline scores for the BPI20 datasets are explained by their relatively low number of activity classes and highly skewed prefix distributions  (see Table~\ref{datasets}). This leaves little room for learned models to improve upon our simple argmax model.

\vspace{-1mm}\paragraph{Why Goliath stumbled:}

Our research suggests that LLMs' pretrained priors are not effectively leveraged for NAP, as fine-tuning alone does not resolve the vocabulary mismatch. Moreover, the relatively low business process complexity (see Table~\ref{datasets}) does not appear to warrant larger and more complex architectures. Instead, comparatively small models trained from scratch can fit the target distribution directly, unencumbered by mismatched priors.

\vspace{-1mm}\paragraph{Implications:}
Our research has three major implications. First, researchers and practitioners alike should first consider a simple and explainable baseline model like our argmax. None of the NAP papers we cited included such a simple baseline model, creating uncertainty about the relevance of the proposed techniques, models, and methods. Second, when the baseline proves insufficient and a more powerful model is warranted, priority should be given to training small Transformer models from scratch, as our results suggest that this is more beneficial than increasing model size or relying on pretrained or distilled architectures. Third, for datasets with high activity diversity or limited training data, uncertainty quantification \cite{weytjens2022learning} can improve practical utility by  trading coverage for accuracy.

\vspace{-1mm}\paragraph{Limitations:}
The validity and generalizibility of our results is limited by our selection of pretrained models from the GPT-2 and Qwen-3 families only, none larger than 4B. Training from scratch and distillation was only performed on one architecture (nanoGPT). Only part of the hyperparameter space was explored. Underperformance of distilled models may partly reflect limitations of logits-only distillation rather than the absence of transferable pretraining knowledge. On the data side, only 7 datasets were used, all drawn from the BPI challenge logs commonly used in the PPM literature; broader coverage across domains and process types would improve the generalizability of our findings. The datasets may not contain all relevant explanatory variables. Process interdependencies were ignored.

\vspace{-1mm}\paragraph{Future research:}We suggest extending this work to other PPM tasks, potentially in a multitask setting, and considering process interdependencies. Transfer learning \cite{berti2026context} with alternative generic representations for different datasets could improve results. Enhancing our sequence modeling approach with symbolic AI techniques or combining it with classical process mining techniques present further research directions. Stratifying by prefix length could reveal whether learned models gain an advantage over argmax as context grows.

\vspace{-1mm}
\section{Conclusion}\label{sec:8-Conclusion}
\vspace{-1mm}
David held his own. On most datasets, a count-based argmax model — conditioning only on the last activity — performs close to fine-tuned LLMs up to eight orders of magnitude larger, despite the latter having access to the complete prefix. However, comparatively small Transformer models trained from scratch often outperform both the argmax baseline and fine-tuned LLMs. Only on BPI12 and BPI17 do Transformer models clearly justify the additional complexity. The broader message for researchers and practitioners is clear: do not bring a giant to a fight where a sling will do. 
The PPM field's tendency to benchmark against other DL models, rather than against simple baselines, has obscured how much of the apparent progress is of practical relevance. Furthermore, smaller Transformer models trained from scratch consistently match or outperform fine-tuned LLMs, suggesting that generic language priors learned during pretraining do not transfer to NAP via vocabulary-adapted direct sequence modeling.

Rather than discouraging further research, these findings establish a clear starting point for the community. Together with dataset benchmarks and standardized preprocessing for NAP \cite{10.1007/978-3-030-94343-1_2}, the findings provide the infrastructure against which genuine progress can be measured. With the bar now clearly set, future work can move beyond architectural scaling and focus on what actually drives performance---whether through richer process-aware representations, transfer learning, multi-task objectives, or entirely new modeling paradigms.

\paragraph{Note on AI use:} We used AI agents, based on GPT-5.2 (later 5.4) and Sonnet-4.6 for coding, visualization, and language editing assistance. \textit{Credits:} We thank Daniel Emrani for contributing to earlier implementation work.
\vspace{-2mm}

\renewcommand{\doi}[1]{}

\bibliographystyle{splncs04}
\bibliography{references}

@misc{oyamada2025domainadaptationllmsprocess,
      title={{Domain Adaptation of LLMs for Process Data}}, 
      author={Rafael Seidi Oyamada and Jari Peeperkorn and Jochen De Weerdt and Johannes De Smedt},
      year={2025},
      eprint={2509.03161},
      archivePrefix={arXiv},
      primaryClass={cs.CL},
      url={https://arxiv.org/abs/2509.03161}, 
}

@InProceedings{10.1007/978-3-031-78666-2_15,
author="Brennig, Katharina
and Kaltenpoth, Sascha
and M{\"u}ller, Oliver",
title="{Straight Outta Logs: Can Large Language Models Overcome Preprocessing in Next Event Prediction?}",
booktitle="BPM Workshops",
year="2025",
publisher="Springer Nature Switzerland",
address="Cham",
pages="197--208",
isbn="978-3-031-78666-2"
}

@InProceedings{10.1007/978-3-030-94343-1_2,
author="Weytjens, Hans
and De Weerdt, Jochen",
title="{Creating Unbiased Public Benchmark Datasets with Data Leakage Prevention for Predictive Process Monitoring}",
booktitle="BPM Workshops",
year="2022",
publisher="Springer International Publishing",
address="Cham",
pages="18--29",
isbn="978-3-030-94343-1"
}

@inproceedings{wuyts2024sutran,
  title={{Sutran: an Encoder-Decoder Transformer for Full-Context-Aware Suffix Prediction of Business Processes}},
  author={Wuyts, Brecht and Vanden Broucke, Seppe and De Weerdt, Jochen},
  booktitle={2024 6th ICPM},
  pages={17--24},
  year={2024},
  organization={IEEE}
}

@inproceedings{rebmann2024evaluating,
  title={{Evaluating the Ability of LLMs to Solve Semantics-Aware Process Mining Tasks}},
  author={Rebmann, Adrian and Schmidt, Fabian David and Glava{\v{s}}, Goran and van Der Aa, Han},
  booktitle={2024 6th ICPM},
  pages={9--16},
  year={2024},
  organization={IEEE}
}

@article{weytjens2022learning,
  title={{Learning Uncertainty with Artificial Neural Networks for Predictive Process Monitoring}},
  author={Weytjens, Hans and De Weerdt, Jochen},
  journal={Applied Soft Computing},
  volume={125},
  pages={109134},
  year={2022},
  publisher={Elsevier}
}

@InProceedings{10.1007/978-3-319-59536-8_30,
author="Tax, Niek
and Verenich, Ilya
and La Rosa, Marcello
and Dumas, Marlon",
title="{Predictive Business Process Monitoring with LSTM Neural Networks}",
booktitle="Advanced Information Systems Engineering",
year="2017",
publisher="Springer International Publishing",
address="Cham",
pages="477--492",
isbn="978-3-319-59536-8"
}

@misc{bukhsh2021processtransformerpredictivebusinessprocess,
      title={{ProcessTransformer: Predictive Business Process Monitoring with Transformer Network}}, 
      author={Zaharah A. Bukhsh and Aaqib Saeed and Remco M. Dijkman},
      year={2021},
      eprint={2104.00721},
      archivePrefix={arXiv},
      primaryClass={cs.LG},
      url={https://arxiv.org/abs/2104.00721}, 
}

@article{CASCIANI2026102642,
title = {{Enhancing Next Activity Prediction in Process Mining with Retrieval-Augmented Generation}},
journal = {Information Systems},
volume = {137},
pages = {102642},
year = {2026},
issn = {0306-4379},
doi = {10.1016/j.is.2025.102642},
curl = {https://www.sciencedirect.com/science/article/pii/S0306437925001280},
author = {Angelo Casciani and Mario Luca Bernardi and Marta Cimitile and Andrea Marrella}
}

@inproceedings{pasquadibisceglie2024lupin,
  title={{Lupin: A LLM Approach for Activity Suffix Prediction in Business Process Event Logs}},
  author={Pasquadibisceglie, Vincenzo and Appice, Annalisa and Malerba, Donato},
  booktitle={2024 6th ICPM},
  pages={1--8},
  year={2024},
  organization={IEEE}
}

@article{vaswani2017attention,
  title={{Attention is All You Need}},
  author={Vaswani, Ashish and Shazeer, Noam and Parmar, Niki and Uszkoreit, Jakob and Jones, Llion and Gomez, Aidan N and Kaiser, {\L}ukasz and Polosukhin, Illia},
  journal={Advances in Neural Information Processing Systems},
  volume={30},
  year={2017}
}

@article{radford2018improving,
  title={{Improving Language Understanding by Generative Pre-Training}},
  author={Radford, Alec and Narasimhan, Karthik and Salimans, Tim and Sutskever, Ilya and others},
  year={2018},
  publisher={San Francisco, CA, USA}
}

@article{nguyen2025next,
  title={{Next Process-Activity Prediction using Switch-Transformer: Approach, Visualization, and Performance Evaluation}},
  author={Nguyen, Thanh-Hai and Pham, Dinh-Lam and Kim, Kwanghoon Pio},
  journal={Knowledge and Information Systems},
  volume={67},
  number={12},
  pages={11827--11854},
  year={2025},
  publisher={Springer}
}

@InProceedings{10.1007/978-3-319-98648-7_27,
author="Di Francescomarino, Chiara
and Ghidini, Chiara
and Maggi, Fabrizio Maria
and Milani, Fredrik",
title="{Predictive Process Monitoring Methods: Which One Suits Me Best?}",
booktitle="International Conference on BPM",
year="2018",
publisher="Springer International Publishing",
address="Cham",
pages="462--479",
isbn="978-3-319-98648-7"
}

@inproceedings{de2021good,
  title={{As Good as New. How to Successfully Recycle English GPT-2 to Make Models for Other Languages}},
  author={de Vries, Wietse and Nissim, Malvina},
  booktitle={Findings of the Association for Computational Linguistics: ACL-IJCNLP 2021},
  pages={836--846},
  year={2021}
}

@article{rama2021deep,
  title={{Deep Learning for Predictive Business Process Monitoring: Review and Benchmark}},
  author={Rama-Maneiro, Efr{\'e}n and Vidal, Juan C and Lama, Manuel},
  journal={IEEE Transactions on Services Computing},
  volume={16},
  number={1},
  pages={739--756},
  year={2021},
  publisher={IEEE}
}

@inproceedings{van2023experiment,
  title={{An Experiment on Transfer Learning for Suffix Prediction on Event Logs}},
  author={van Luijken, Mathieu and Ketyk{\'o}, Istv{\'a}n and Mannhardt, Felix},
  booktitle={International Conference on BPM},
  pages={31--43},
  year={2023},
  organization={Springer}
}

@inproceedings{pyrih2025llms,
  title={{LLMs that Understand Processes: Instruction-tuning for Semantics-Aware Process Mining}},
  author={Pyrih, Vira and Rebmann, Adrian and van der Aa, Han},
  booktitle={2025 7th ICPM},
  pages={1--8},
  year={2025},
  organization={IEEE}
}

@article{EVERMANN2017129,
title = {{Predicting Process Behaviour using Deep Learning}},
journal = {Decision Support Systems},
volume = {100},
pages = {129-140},
year = {2017},
issn = {0167-9236},
doi = {10.1016/j.dss.2017.04.003},
curl = {https://www.sciencedirect.com/science/article/pii/S0167923617300635},
author = {Joerg Evermann and Jana-Rebecca Rehse and Peter Fettke}
}

@article{10.1162/neco.1997.9.8.1735,
author = {Hochreiter, Sepp and Schmidhuber, J\"{u}rgen},
title = {{Long Short-Term Memory}},
year = {1997},
issue_date = {November 15, 1997},
publisher = {MIT Press},
address = {Cambridge, MA, USA},
volume = {9},
number = {8},
issn = {0899-7667},
curl = {https://doi.org/10.1162/neco.1997.9.8.1735},
doi = {10.1162/neco.1997.9.8.1735},
journal = {Neural Comput.},
month = nov,
pages = {1735–1780},
numpages = {46}
}

@article{brown2020language,
  title={{Language Models are Few-Shot Learners}},
  author={Brown, Tom and Mann, Benjamin and Ryder, Nick and Subbiah, Melanie and Kaplan, Jared D and Dhariwal, Prafulla and Neelakantan, Arvind and Shyam, Pranav and Sastry, Girish and Askell, Amanda and others},
  journal={Advances in Neural Information Processing Systems},
  volume={33},
  pages={1877--1901},
  year={2020}
}

@article{hinton2015distilling,
  title={{Distilling the Knowledge in a Neural Network}},
  author={Hinton, Geoffrey and Vinyals, Oriol and Dean, Jeff},
  journal={arXiv preprint arXiv:1503.02531},
  year={2015}
}

@article{han2024parameter,
  title={{Parameter-Efficient Fine-Tuning for Large Models: A Comprehensive Survey}},
  author={Han, Zeyu and Gao, Chao and Liu, Jinyang and Zhang, Jeff and Zhang, Sai Qian},
  journal={arXiv preprint arXiv:2403.14608},
  year={2024}
}

@inproceedings{press2017using,
  title={{Using the Output Embedding to Improve Language Models}},
  author={Press, Ofir and Wolf, Lior},
  booktitle={Proceedings of the 15th Conference of the European Chapter of the Association for Computational Linguistics: Volume 2, Short Papers},
  pages={157--163},
  year={2017}
}

@article{TAMA2019233,
title = {{An Empirical Comparison of Classification Techniques for Next Event Prediction using Business Process Event Logs}},
journal = {Expert Systems with Applications},
volume = {129},
pages = {233-245},
year = {2019},
issn = {0957-4174},
doi = {10.1016/j.eswa.2019.04.016},
curl = {https://www.sciencedirect.com/science/article/pii/S0957417419302465},
author = {Bayu Adhi Tama and Marco Comuzzi}
}

@misc{weinzierl2020empiricalcomparisondeepneuralnetworkarchitectures,
      title={{An Empirical Comparison of Deep-Neural-Network Architectures for Next Activity Prediction using Context-Enriched Process Event Logs}}, 
      author={S. Weinzierl and S. Zilker and J. Brunk and K. Revoredo and A. Nguyen and M. Matzner and J. Becker and B. Eskofier},
      year={2020},
      eprint={2005.01194},
      archivePrefix={arXiv},
      primaryClass={cs.LG},
      url={https://arxiv.org/abs/2005.01194}, 
}

@Article{app151910434,
AUTHOR = {Hwang, Taewook and Seo, Hyein and Jung, Jeesu and Jung, Sangkeun},
TITLE = {Exploring Selective Layer Freezing Strategies in Transformer Fine-Tuning: NLI Classifiers with Sub-3B Parameter Models},
JOURNAL = {Applied Sciences},
VOLUME = {15},
YEAR = {2025},
NUMBER = {19},
ARTICLE-NUMBER = {10434},
URL = {https://www.mdpi.com/2076-3417/15/19/10434},
ISSN = {2076-3417},
DOI = {10.3390/app151910434}
}

@article{berti2026context,
  title={{An In-Context Foundation Model for Predictive Process Monitoring on Event Logs}},
  author={Berti, Alessandro and Van Der Aalst, Wil MP},
  journal={IEEE Access},
  year={2026},
  publisher={IEEE}
}

\end{document}